\documentclass[10pt,conference]{IEEEtran} 

\usepackage{graphicx}
\usepackage[utf8]{inputenc}
\usepackage{graphicx}     
\usepackage{amsmath,amssymb}
\usepackage{booktabs}     
\usepackage{hyperref}     
\usepackage{cite}         
\usepackage{xcolor}       
\usepackage{fancyhdr}

\title{WAVE-DETR Multi-Modal Visible and Acoustic Real-Life Drone Detector}

\author{
  Razvan Stefanescu\textsuperscript{1}, Ethan Oh\textsuperscript{1}, Ruben Vazquez\textsuperscript{1}, 
  Chris Mesterharm \textsuperscript{1}, Constantin Serban\textsuperscript{1}, Ritu Chadha\textsuperscript{1} \\
  \textit{\textsuperscript{1}Peraton Labs, Basking Ridge, USA} \\
  \texttt{razvanstefanescu@peraton.com (Corresponding author)}
}

\begin{document}

\maketitle

\begin{abstract}
We introduce a multi-modal WAVE-DETR drone detector combining visible RGB and acoustic signals for robust real-life UAV object detection. Our approach fuses visual and acoustic features in a unified object detector model relying on the Deformable DETR and Wav2Vec2 architectures, achieving strong performance under challenging environmental conditions. Our work leverage the existing Drone-vs-Bird dataset and the newly generated ARDrone dataset containing more than $7,500$ synchronized images and audio segments. We show how the acoustic information is used to improve the performance of the Deformable DETR object detector on the real ARDrone dataset. We developed, trained and tested four different fusion configurations based on a gated mechanism, linear layer, MLP and cross attention. The Wav2Vec2 acoustic embeddings are fused with the multi resolution feature mappings of the Deformable DETR and enhance the object detection performance over all drones dimensions.   
The best performer is the gated fusion approach, which improves the mAP of the Deformable DETR object detector on our in‑distribution and out‑of‑distribution ARDrone datasets by $11.1\%$ to $15.3\%$ for small drones across all IoU thresholds between $0.5$ and $0.9$. The mAP scores for medium and large drones are also enhanced, with overall gains across all drone sizes ranging from $3.27\%$ to $5.84\%$.
\end{abstract}

\section{Introduction}

Object detection is a classic and fundamental task in computer vision \cite{jiang2022review, ren2015faster, carion2020end, zhu2020deformable}. Its objective consists in locating and classifying objects within an image by drawing bounding boxes around them, making it fundamental to numerous real-world applications. With the advancement of automotive technology, object detection models have become integral components of autonomous driving systems \cite{caesar2020nuscenes, bijelic2020seeing}. Within the end-to-end self-driving pipeline \cite{hu2022st}, object detection plays a critical role in perception tasks such as identifying vehicles, pedestrians, road signs, cyclists, and lane markings. These detections feed into higher-level modules like tracking \cite{liu2021robust}, path planning \cite{paden2016survey}, and control systems \cite{caporale2018planning}, enabling vehicles to make safe navigation decisions in dynamic environments. Safety is paramount in autonomous driving, and conditions such as poor visibility at night or adverse weather significantly degrade the performance of camera-only detection systems. To address this, multimodal object detection \cite{chen2023futr3d, chu2023mt} combining RGB cameras with lidar, radar, and audio sensors is increasingly being adopted to provide more robust and reliable detection under all conditions.

Beyond autonomous driving, object detection has become essential for security, defense and military applications, particularly in detecting and tracking unmanned aerial vehicles (UAVs) or drones \cite{bhattacharya2025drone}. As drones become more prevalent in both civilian and military contexts, they pose potential risks related to privacy breaches, smuggling, and even targeted attacks. Object detection algorithms are employed in surveillance systems to identify UAVs in complex environments \cite{zhai2023yolo, kyrkou2018dronet,  du2018unmanned, liu2024yolc, coluccia2024}, often in combination with radar, infrared, RF or acoustic sensors. These systems can be used to trigger alerts, activate countermeasures, or guide tracking systems for drone mitigation. Accurate and real-time detection of UAVs is therefore critical for maintaining airspace security and protecting sensitive infrastructure.

In this work, we present a novel multi-modal WAVE-DETR drone detector by fusing visible RGB and acoustic signals for robust real-life UAV object detection. We also generate a new acoustic and RGB ARDrone dataset, consisting of 77 video sequences and over $7,500$ synchronized RGB–audio pairs, sampled at a frequency of one second. Our architecture is specially designed to improve the detection of small UAVs. The architecture is built on the shoulders of the Deformable DETR model \cite{zhu2020deformable, carion2020end} which combines traditional Resnet backbones \cite{he2016deep} for feature mapping extraction with an encoder-decoder transformer that refines the query points/bounding boxes and classification scores during training. Since the multi-resolution features maps are known for their capability to improve the detection of small UAV, we introduce a fusion layer inside the Deformable DETR before the encoder. It mixes the RGB features maps with audio embeddings extracted from the wav2vec2 foundational model backbone \cite{baevski2020wav2vec}  just before its transformer component. The audio feature is carefully aligned with each of the RGB feature maps and then four different fusion configurations are applied based on a linear layer, a multi-layer perceptron, a gating mechanism and cross-attention. 

To achieve better performances than the RGB Deformable DETR model, we fine-tuned single modal RGB and acoustic architectures as follows. In the case of the RGB Deformable DETR, we start from a pre-trained COCO version \cite{lin2014coco} and then fine tuned on the ARDrone dataset whereas the wav2vec2 model is pre-trained on Librispeech corpus and TIMIT datasets and fine-tuned on the Drone-detection \cite{svanstrom2021real} and ARDrone datasets. The ablation experiments testing the various fusion layers and dropout and learning rates reveal that the gating fusion mechanism is the best performer for all drone size categories.                   
Here are the novelties of our paper:

1. Generated a new multi-modal audio and RGB ARDrone dataset with more than $7,500$ annotated pairs.

2. Introduced a novel multi-modal WAVE-DETR acoustic and RGB object detector with four different fusion layers combining wav2vec2 and Deformable DETR architectures. The fusion layer and acoustic features provide excellent improvements when compared with the RGB Deformable DETR architecture on small drones. 

The paper is structured as follows. Section II discusses the existing literature related to our work. Section III covers existing drone datasets including our newly generated ARDrone dataset. The WAVE-DETR architecture and fusion layers are presented in details in Section IV. The ablation experiments are provided in Section V followed by the Conclusion section. 

\section{Related Work}
The most popular solution for the multi-modal drone detection problem utilizes a combination of deep learning unimodal models followed by traditional Kalman Filter \cite{kalman1960, liu2022enkf} fusion relying on a drone pose state model. For example, Svanstr{\"o}m et al. \cite{svanstrom2021real} presents a multimodal drone detection system using RGB, thermal, and acoustic data, where detection is formulated as a classification task (drone vs. non-drone) without bounding box prediction. YOLOv2 \cite{jiang2022review} is used for the visual modalities, while Mel-Frequency Cepstral Coefficient (MFCC) \cite{sidhu2025mfcc} features combined with an LSTM network are employed for audio-based classification. The acoustic detection emerged as the most effective method. MUTES \cite{ding2023drone} is another drone detection and tracking system that integrates multiple sensor modules, including a microphone array, camera, and LiDAR. In the first stage, an acoustic denoising and source localization process—based on beamforming and a deep learning architecture (1D CNN and LSTM)—is used to estimate an acoustic map and to distinguish relevant drone signals from environmental noise. In the second stage, unimodal estimates of azimuth, elevation, and radial distance are fused using a Kalman filter with a constant-velocity ballistic motion model to track the drone’s position.

Recent advances in multimodal object detection have introduced robust end-to-end architectures that effectively integrate heterogeneous sensor data. MT-DETR \cite{chu2023mt} proposes a multistage detection framework that incorporates fusion and enhancement modules along with a hierarchical fusion mechanism to jointly leverage camera, LiDAR, and radar data. Similarly, MDETR \cite{kamath2021mdetr} introduces a transformer-based architecture for modulated detection that performs end-to-end object detection conditioned on raw text queries such as captions or questions, enabling applications beyond traditional detection, including few-shot learning and visual question answering. Complementing these approaches, FUTR3D \cite{chen2023futr3d} presents a unified sensor fusion framework for 3D detection, where modality-specific features are first encoded individually and then fused through a Modality-Agnostic Feature Sampler (MAFS) in a unified space. A transformer decoder subsequently processes 3D queries to output object predictions. Together, these models represent a significant step toward flexible, scalable, and context-aware multimodal detection systems. Our proposed WAVE-DETR architecture belongs to the category of end-to-end detectors.


\section{Drone Datasets}
\label{sec:formatting}
Among the most diverse and widely used drone detection datasets are the Drone-vs-Bird \cite{coluccia2024} and MMAUD \cite{yuan2024mmaud} datasets. The Drone-vs-Bird Detection Challenge dataset comprises 77 video sequences captured using both static and moving cameras, with resolutions ranging from 720×576 to 3840×2160 pixels. While it offers a rich variety of visual scenarios for detecting and distinguishing drones from birds, it is limited to the RGB modality, as audio data is not provided. In contrast, MMAUD is a multimodal dataset that spans approximately 1,700 seconds of recordings, incorporating a wide array of sensory inputs including stereo vision, multiple LiDARs, radars, and microphone arrays. It is specifically designed to address the challenges of detecting small UAVs in complex environments by leveraging complementary data modalities.

{\it ARDrone dataset:} For this study, we generated
a multi-modal real drone dataset using a DJI Mavic 3 Pro drone flown in a variety of environmental and acoustic conditions. The sensing system comprises a Google Pixel 2 smartphone equipped with a built-in camera and microphone, which was mounted on a tripod to ensure stability during data collection. The data was collected as a video of $1920$ x $1080$ resolution and processed into RGB frames and audio segments.

Our data collection protocol integrated as much variability as possible in both vision and acoustic spaces. To account for light variability, we collected drone frames in diverse cloud conditions (clear sky, scattered clouds, overcast, etc.) and at different times of the day, including evening. We aimed to capture variations in viewing angles at various distances by rotating the in-flight drone around its vertical (z) axis. This was done at 10, 20 and 30 feet elevations and at every 10 to 15 feet distance from the sensing system up to 60 feet away. Furthermore, the angle of the sensor mounting onto the tripod was varied to account for top-down and bottom-up views where possible. The drone was also captured flying freely in a user-controlled, nonspecific flight-paths of up to 250 feet away from the sensors. As stated earlier, the acoustic data was collected simultaneously and therefore share the same structured variations. 

We took care to capture the drone in different settings to cover a large diversity of backgrounds. Our dataset includes the drone flying among open fields, trees, cars, buildings and bodies of water. Acoustically, we were mindful of capturing variations in environmental noises (the drone noise reflection when flying near a large surface). Steps were taken to include background sounds such as wind, birds, airplanes, cars, human utterances, footsteps and dogs barking. 

\begin{figure}[t]
  \centering
  \includegraphics[width=1\linewidth]{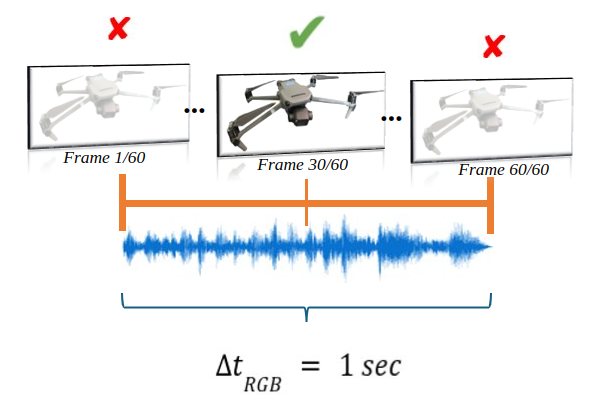}
  \caption{Time synchronization between RGB frames and audio segments is achieved by selecting the RGB frame that occurs at the midpoint of each 1-second audio segment.}
  \label{fig:Time_sync}
\end{figure}

Since the smartphone hardware already solved the time syncronization between the RGB frames and acoustic sensors, we only had to generate pairs between the two modalities outputs. The videos were  discretized in RGB frames at a rate of $60$ fps and the continuous audio stream  was split into $1$s audio segments. The ARdrone samples are formed with $1$s audio segments and RGB frames aligned with the center of the audio segments as seen in Figure \ref{fig:Time_sync}. 

For object detection annotation, we employed the annotation application CVAT to label the bounding boxes and adhered to COCO format \cite{lin2014coco}. We automated the process by using a Deformable DETR detector trained on the Drone-vs-Birds dataset \cite{coluccia2024}. We randomly selected a subset of the $56$k images to fine-tune the Deformable DETR detector, which was obtained from HuggingFace. 

This setup allowed us to generate $7,699$ RGB and audio segments pairs, adhering to the two class COCO format: drone and background classes. COCO's JSON formatting allowed us to inject additional meta-data into the annotation and the distances between the sensors and the drone were recorded along with the bounding box coordinates. We also augmented the drone audio segments with samples from drone detection dataset \cite{svanstrom2021real} containing IR, visible and audio modalities.  

\begin{figure}[t]
  \centering
  \includegraphics[width=1\linewidth]{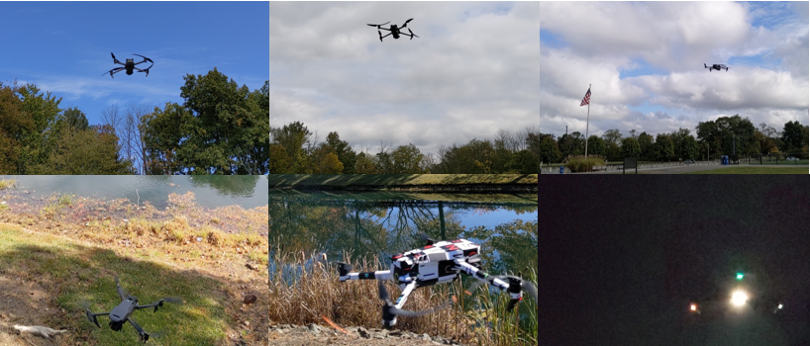}

  \caption{Sample frames extracted from the Peraton dataset videos showing the variability of the dataset.}
  \label{fig:Peaton_dataset}
\end{figure}

Some examples of the different drone orientations with variable backgrounds and light conditions can be observed in Figure \ref{fig:Peaton_dataset}. The RGB sensed drone areas ranged from $9$ square pixels to a maximum of $535$k square pixels depending on the distance between the drone and sensors. More than $2$k images included large drones ( $>96^2$ square pixels) whereas small drones ($<32^2$ square pixels) were present in $3$k images. The resolution degradation of small-sized drones poses a great challenge in detection tasks. Our dataset's distribution of the RGB-sensed drone bounding boxes dimensions is illustrated in Figure \ref{fig:object_area_distribution}.

\begin{figure}[t]
  \centering
  \includegraphics[width=1\linewidth]{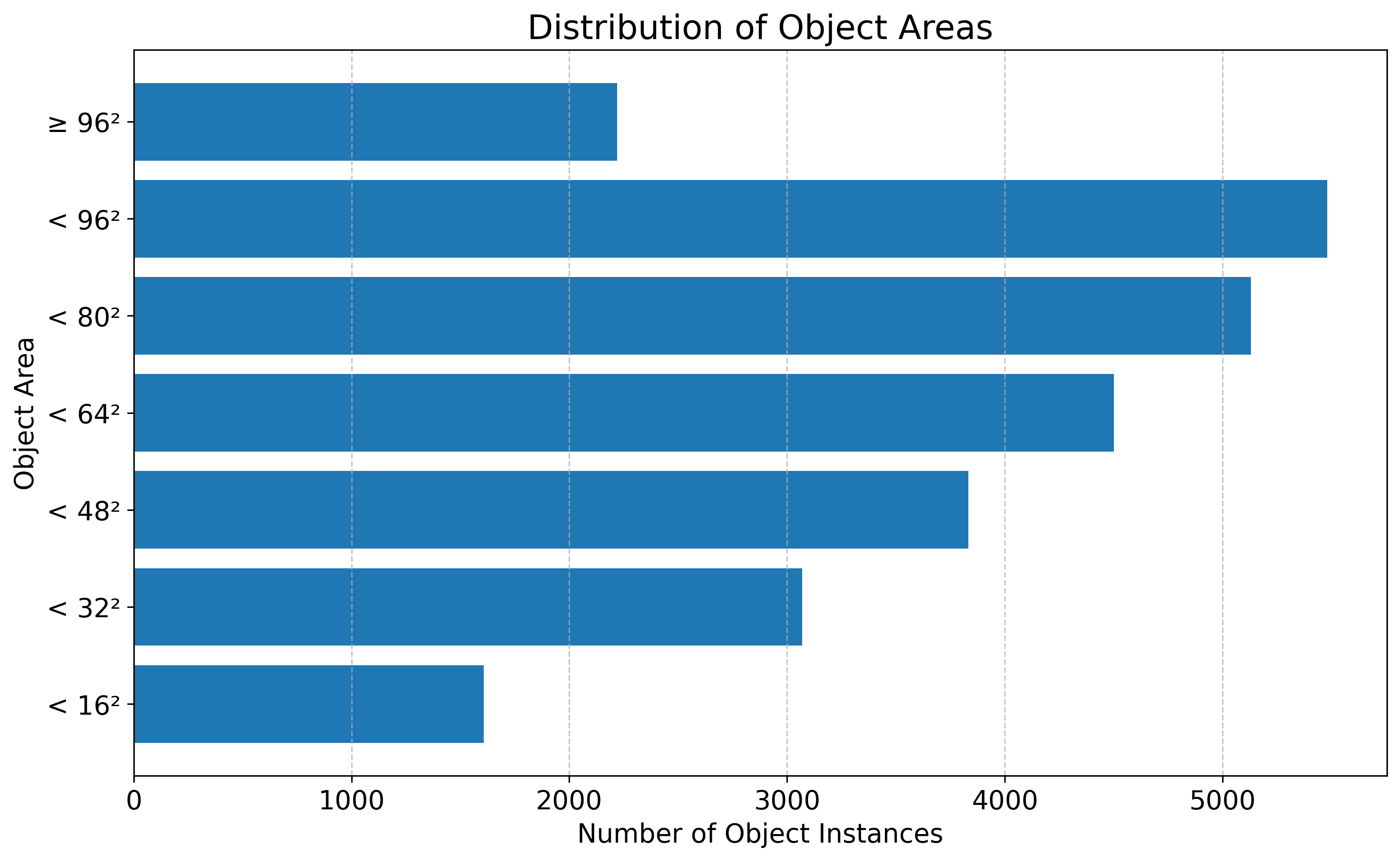}

  \caption{Distribution of drone sizes across the ground truth annotations in the entire dataset.}
  \label{fig:object_area_distribution}
\end{figure}

\section{Multi-Modal audio and RGB Deformable DETR architecture}

Our drone detection architecture extends the Deformable DETR model with fusion layers and combines the audio features resulting from the wav2vec2 backbone with the RGB feature maps of the Deformable DETR. Before describing the multi-modal architecture, we present some details of the DETR object detector and wav2vec2 models.   

\subsection{DETR object detectors class}

DETR object detector \cite{carion2020end} has been introduced to eliminate the need of using many hand-crafted components such as anchor generation, rule-based training target assignment and non-maximum suppression post-procession. From an engineering point of view, it is the first fully end-to-end object detector. It relies on the Hungarian algorithm for bipartite matching between predicted and ground-truth objects in contrast to the traditional anchor-based methods like Faster R-CNN \cite{girshick2015fast} and YOLO \cite{jiang2022review}. The matching step is purely a combinatorial optimization process involving discrete operations and it is performed before the loss function is applied. Unmatched predictions are treated as background while unmatched ground-truths are considered missed detections. The DETR architecture combines an ImageNet pre-trained ResNet model with an encoder-decoder transformer \cite{vaswani2017attention} to guide the query points to the actual ground-truth bounding bounds center coordinates. Despite its state-of-the-art architecture, DETR suffers from slow convergence \cite{zhu2020deformable} and on the COCO benchmark, DETR requires $500$ epochs to converge which is about $10-20$ slower than Faster R-CNN. Additionally, DETR performance at detecting small objects is relatively low in comparison with the  object detectors that usually exploit multi-scale features and multi-resolution feature maps. 

\begin{figure}[t]
  \centering
  \includegraphics[width=1\linewidth]{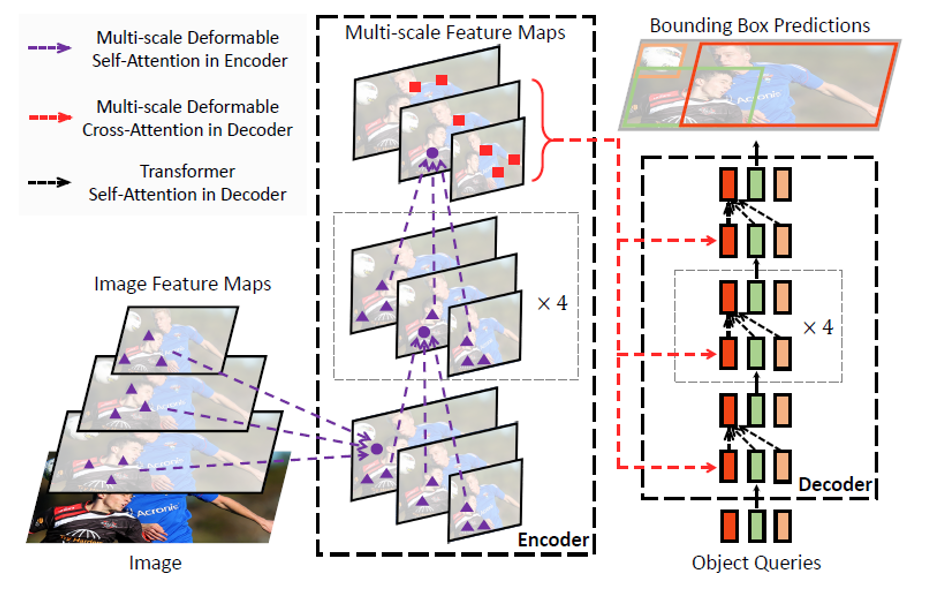}

  \caption{Deformable DETR Architecture with its 4 different feature mAPs. The image was taken from \cite{zhu2020deformable}}
  \label{fig:Deformable_Detr}
\end{figure}

Deformable DETR \cite{zhu2020deformable} improves the convergence rate of DETR by implementing an efficient attention mechanism for processing image feature maps. The sparse attention mechanism attends to a small set of sampling locations to select important key elements out of all four feature mAPs at various spatial resolutions. It aggregates multi-scale features as an alternative to the FPN \cite{lin2017feature}. The Deformable DETR architecture is described in Figure \ref{fig:Deformable_Detr} and we employed it as our stand alone RGB detector (baseline) as well as the platform for the multi-modal WAVE-DETR object detector.  

\subsection{Wave2Vec2 Audio Foundational Model}

\begin{figure}[t]
  \centering
  \includegraphics[width=1\linewidth]{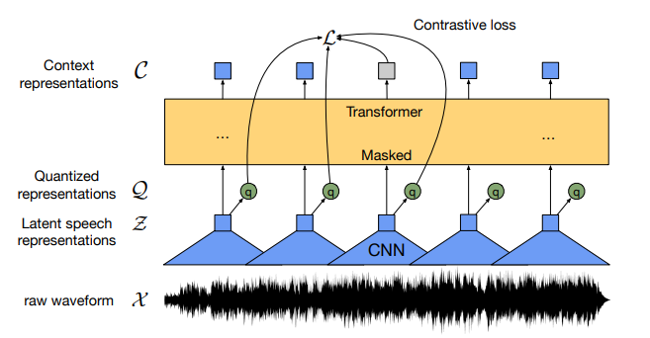}

  \caption{Self Supervised wav2vec2 architecture. Image is reproduced from \cite{baevski2020wav2vec20frameworkselfsupervised}}
  \label{fig:object_area_distribution}
\end{figure}

The Wav2Vec2 model is a foundational model that processes audio sequences for downstream applications.  \cite{baevski2020wav2vec20frameworkselfsupervised}. Wav2Vec2 uses a self-supervised learning approach where general data representations for audio samples are learned from unlabeled data during training and its architecture is depicted in Figure \ref{fig:object_area_distribution}. Wav2Vec2 is trained by feeding raw audio waveforms to a Convolutional Neural Network to generate latent representations of the waveforms. The latent representations of the waveforms are then fed to a Transformer to generate the output context representations, leveraging self-attention to capture dependencies of the latent representations. The context representations are then compared to the quantized versions of the latent representations via a contrastive loss for training the model.

On downsteam tasks, Wav2Vec2 is able to compete with approaches with significantly less data. In this paper, we are using Wav2Vec2 as a standalone classifier without bounding boxes predictions and integrate its Convolutional Neural Network backbone into the Wave-DETR object detector.

{\subsection{The multi-modal WAVE-DETR fusion architecture}}

\begin{table*}[h!]
\centering
\begin{tabular}{lcccc}
\hline
\textbf{Method} & \textbf{No dropout} & \textbf{Dropout 0.1} & \textbf{Dropout 0.2} & \textbf{Dropout 0.3} \\
\hline
COCO-init RGB & 0.702 & -- & -- & -- \\ 
Local-Init Linear Fusion & 0.717 & 0.716 & 0.704 & 0.703 \\
COCO-init Linear Fusion & 0.693 & 0.707 & 0.702 & 0.697 \\
Local-Init Gated Fusion & \textbf{0.725} & 0.724 & 0.716 & 0.716 \\
COCO-init Gated Fusion & 0.703 & 0.703 & 0.695 & 0.708 \\
Local-Init MLP Fusion & 0.702 & 0.712 & 0.707 & 0.699 \\
COCO-init MLP Fusion & 0.701 & 0.688 & 0.711 & 0.694 \\
Local-Init Cross Attention Fusion & 0.692 & 0.681 & 0.701 & 0.676 \\
COCO-init Cross Attention Fusion & 0.673 & 0.664 & 0.687 & 0.679 \\
\hline
\end{tabular}
\caption{In distribution mAP for all small, medium, and large objects across IoU thresholds between 0.5 and 0.9.}
\label{tab:in_distribution_results}
\end{table*}

\begin{table*}[h!]
\centering
\begin{tabular}{lcccc}
\hline
\textbf{Method} & \textbf{No dropout} & \textbf{Dropout 0.1} & \textbf{Dropout 0.2} & \textbf{Dropout 0.3} \\
\hline
COCO-init RGB & 0.486 & -- & -- & -- \\
Local-Init  Linear Fusion & 0.506 & 0.512 & 0.506 & 0.508 \\
COCO-init Linear Fusion & 0.481 & 0.510 & 0.510 & 0.484 \\
Local-Init  Gated Fusion & 0.521 & \textbf{0.54} & 0.506 & 0.519 \\
COCO-init Gated Fusion & 0.497 & 0.491 & 0.490 & 0.488 \\
Local-Init  MLP Fusion & 0.513 & 0.512 & 0.490 & 0.503 \\
COCO-init MLP Fusion & 0.497 & 0.482 & 0.507 & 0.498 \\
Local-Init Cross Attention Fusion & 0.489 & 0.487 & 0.519 & 0.481 \\
COCO-init Cross Attention Fusion & 0.485 & 0.450 & 0.487 & 0.475 \\
\hline
\end{tabular}
\caption{In distribution mAP for the small objects across IoU thresholds between 0.5 and 0.9.}
\label{tab:small_objects_results}
\end{table*}

\begin{figure}[t]
  \centering
  \includegraphics[width=1\linewidth]{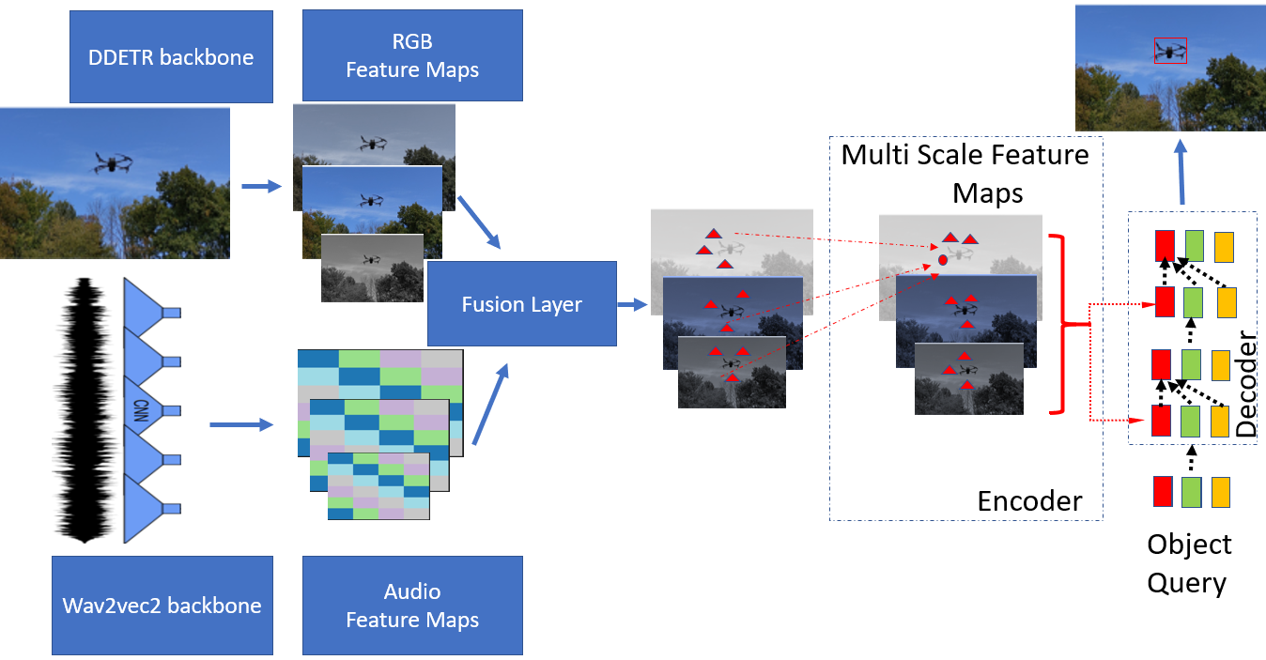}

  \caption{Self Supervised Wav2vec2+ Deformable DETR architecture fusing both RGB and audio modalities to improve object detection. }
  \label{fig:Deformable_Wav2Vec2_DETR}
\end{figure}

It is well known that object detectors struggle to identify small objects. Deformable DETR addresses this limitation by introducing multi‑scale feature maps to improve small‑object detection. We extend Deformable DETR architecture by incorporating acoustic features through a fusion layer that combines the RGB multi‑scale feature maps with audio multi‑scale feature maps before the encoder as seen in Figure \ref{fig:Deformable_Wav2Vec2_DETR}. We implemented four different fusion layers including a linear layer, a multi-layer perceptron, a gating mechanism and cross attention. 

The audio embeddings are generated from the backbone of the Wav2Vec2 network integrated inside the Deformable DETR architecture. The audio embeddings are then interpolated to match the dimensions of the RGB feature maps. First, the audio embeddings are reshaped so that their last dimension matches the last dimension of the RGB feature maps. Next, the reshaped audio features are linearly interpolated to align their sequence length with that of each RGB feature map. Once both the RGB and audio features have the same dimensions, they are concatenated along the last dimension. 

\begin{figure}[t]
  \centering
  \includegraphics[width=1\linewidth]{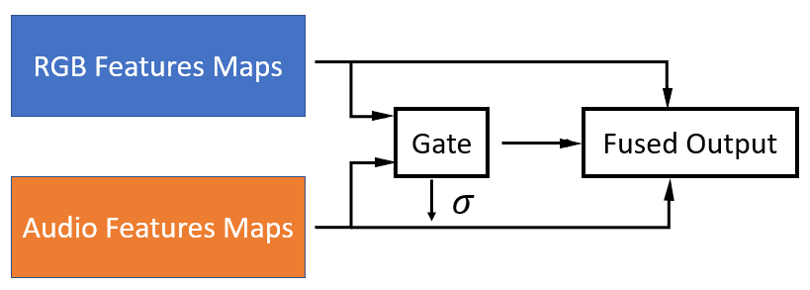}

  \caption{Gate Fusion combining linear and audio feature maps.}
  \label{fig:Gate_fusion_mechanism}
\end{figure}

At this point, each of the fusion layers process the concatenated structure in various ways. The linear fusion applies a linear transformation to reduce the last dimension to the original size of the RGB feature followed by a dropout layer. The multi-layer perceptron extends the fusion layer with a RELU activation function and another dropout layer. The gating mechanism applies a sigmoid function to the output of the linear layer to generate a weighted linear combination between the aligned RGB and audio features as seen in Figure \ref{fig:Gate_fusion_mechanism}. 

In the case of cross‑attention, there is no need to manually align the sequence dimensions of the audio and RGB features. The alignment is handled naturally by the attention mechanism itself, where the RGB features serve as the queries, while the audio embeddings provide the keys and values.
\vskip 0.3cm
\section{Results and discussions}

\subsection{Engineering characteristics}

The training of both the unimodal and multimodal architectures was conducted on a system equipped with four NVIDIA RTX 6000 Ada GPUs each with 48 GB of memory. For the unimodal architectures, training was performed on a single GPU, while we employed PyTorch’s Distributed Data Parallel (DDP) library  and utilized all four GPUs for the multimodal architectures. We increased the batch size and improved convergence rates by implementing the automated mixed precision \cite{micikevicius2018mixed}, gradient checkpointing and accumulation. 

As for the optimization engine, we used the Adam optimizer and applied two learning rate schedulers: ReduceLROnPlateau and CosineAnnealing. In addition, we experimented with different backbone learning rates and varied the dropout parameter within the fusion layers.

\subsection{Datasets distributions}

The ARDrone dataset contains $70$ different videos obtained in various weather conditions (clear sky, light cloud coverage and overcast) with different backgrounds (soccer field, trees, buildings, parking lot, and fence with power lines) and light conditions. For our experiments, we designed two datasets, referred to as in‑distribution and out‑of‑distribution datasets. In the in‑distribution setting, the testing and validation RGB frames and audio segments share the same background, weather and lighting conditions as the training videos. In the out‑of‑distribution setting, the background, weather, and lighting conditions are completely different from those in the training videos.  The in-distribution training, validation and testing datasets contain $4,619$, $1,540$ and $1,540$ image and audio segment pairs while the out-distribution training, validation and testing datasets contain $6,332$, $1,324$, $904$ samples. 

\subsection{Evaluation metrics}

We are employing the mean average precision to evaluate the RGB and multi-modal Deformable DETR detectors, whereas for the Wav2Vec2 classifier, the standard classification evaluation metrics such as precision, recall, F1-score, balanced accuracy and Matthews Correlation Coefficient are applied. The mean average precision is calculated for all drones as well as small, medium and large UAVs across IoUs between 0.5 and 0.9.    

\begin{table*}[h!]
\centering
\begin{tabular}{lcccc}
\hline
\textbf{Method} & \textbf{No dropout} & \textbf{Dropout 0.1} & \textbf{Dropout 0.2} & \textbf{Dropout 0.3} \\
\hline
COCO-Init RGB & 0.765 & -- & -- & -- \\
Local-Init Linear Fusion & 0.797 & 0.794 & 0.781 & 0.769 \\
COCO-Init Linear Fusion & 0.757 & 0.765 & 0.755 & 0.751 \\
Local-Init Gated Fusion & \textbf{0.799} & 0.785 & 0.780 & 0.782 \\
COCO-Init Gated Fusion & 0.771 & 0.768 & 0.755 & 0.780 \\
Local-Init MLP Fusion & 0.757 & 0.780 & 0.789 & 0.762 \\
COCO-Init MLP Fusion & 0.774 & 0.746 & 0.778 & 0.750 \\
Local-Init Cross Attention Fusion & 0.770 & 0.743 & 0.760 & 0.737 \\
COCO-Init Cross Attention Fusion & 0.716 & 0.724 & 0.755 & 0.756 \\
\hline
\end{tabular}
\caption{In distribution mAP for the medium objects across IoU thresholds between 0.5 and 0.9.}
\label{tab:medium_objects_results}
\end{table*}

\begin{table*}[h!]
\centering
\begin{tabular}{lcccc}
\hline
\textbf{Method} & \textbf{No dropout} & \textbf{Dropout 0.1} & \textbf{Dropout 0.2} & \textbf{Dropout 0.3} \\
\hline
COCO-Init RGB & 0.888 & -- & -- & -- \\
Local-Init Linear Fusion & 0.884 & 0.886 & 0.874 & 0.892 \\
COCO-Init Linear Fusion & 0.879 & 0.887 & 0.887 & 0.883 \\
Local-Init Gated Fusion & \textbf{0.906} & 0.894 & 0.897 & 0.894 \\
COCO-Init Gated Fusion & 0.892 & 0.882 & 0.881 & 0.885 \\
Local-Init MLP Fusion & 0.882 & 0.883 & 0.884 & 0.871 \\
COCO-Init MLP Fusion & 0.872 & 0.867 & 0.896 & 0.881 \\
Local-Init Cross Attention Fusion & 0.876 & 0.864 & 0.868 & 0.859 \\
COCO-Init Cross Attention Fusion & 0.866 & 0.852 & 0.859 & 0.853 \\
\hline
\end{tabular}
\caption{In distribution mAP for the large objects across IoU thresholds between 0.5 and 0.9.}
\label{tab:large_objects_results}
\end{table*}

\subsection{Experiments description} 
The baseline for our drone detection problem consists of the RGB Deformable DETR model pretrained on the COCO dataset that was fine tuned using the ARDrone datasets. We also trained a Wav2Vec2 binarry classifier with drone and background classes described in Appendix A. 

The multi-modal architectures consisting of combinations between the wav2vec2 and Deformable DETR models and various fusion layers aim to take advantage of the acoustic information and inter-modalities correlations to improve over the baseline. We vary the RGB Deformable DETR initialization between COCO pretrained and locally fine tuned corresponding to the baseline. The corresponding acoustic weights are only initialized from the locally fine tuned Wav2Vec2 model. 

Using the lightning AI framework for training, our architectures allowed us to test various learning rates in the ResNet-50 backbone of the Deformable DETR and CNN backbone of the Wav2Vec2 components. 
Among the tested learning rates $[0, 1e-4, 1e-5, 1e-6$], the best mAP results for all multi-modal architectures were  obtained for $1e-5$. As such, all the displayed results uses this backbone learning rate. The architectures were trained using $60$ epochs. Our ablation study is completed by testing various dropping rates in the fusion layers. 

\subsection {Naming convention for architectures} The naming convention for the trained architectures starts with RGB Deformable DETR weight initialization. We refer to the single modal RGB Deformable Detr initialized from pretrained COCO and fine tuned on the local dataset as COCO-Init RGB. We utilize fusion in the naming of the multi-modal Wav2Vec2 + RGB Deformable architectures and use the precise fusion layer to refer to it. We refer to the Wav2Vec2 + RGB Deformable Gated fusion architecture with the RGB weights initialized from COCO dataset as Local-Init Gated Fusion.   
\begin{table*}[t!]
\centering
\begin{tabular}{lcccc}
\hline
\textbf{Method} & \textbf{No dropout} & \textbf{Dropout 0.1} & \textbf{Dropout 0.2} & \textbf{Dropout 0.3} \\
\hline
COCO-Init RGB & 0.741 & -- & -- & -- \\
Local-Init Linear Fusion & 0.776 & 0.754 & 0.767 & 0.784 \\
COCO-Init Linear Fusion & 0.713 & 0.751 & 0.734 & 0.759 \\
Local-Init Gated Fusion & 0.777 & \textbf{0.787} & 0.769 & 0.758 \\
COCO-Init Gated Fusion & 0.730 & 0.755 & 0.759 & 0.752 \\
Local-Init MLP Fusion & 0.748 & 0.733 & 0.778 & 0.721 \\
COCO-Init MLP Fusion & 0.732 & 0.703 & 0.736 & 0.720 \\
Local-Init Cross Attention Fusion & 0.689 & 0.682 & 0.701 & 0.607 \\
COCO-Init Cross Attention Fusion & 0.639 & 0.667 & - & - \\ 
\hline
\end{tabular}
\caption{Out of distribution mAP for all the small, medium and large objects and across IoU thresholds between 0.5 and 0.9.}
\label{tab:ood_mAP_results}
\end{table*}

\begin{table*}[t!]
\centering
\begin{tabular}{lcccc}
\hline
\textbf{Method} & \textbf{No dropout} & \textbf{Dropout 0.1} & \textbf{Dropout 0.2} & \textbf{Dropout 0.3} \\
\hline
COCO-Init RGB & 0.443 & -- & -- & -- \\
Local-Init Linear Fusion & 0.483 & 0.461 & 0.503 & 0.496 \\
COCO-Init Linear Fusion & 0.393 & 0.475 & 0.417 & 0.447 \\
Local-Init Gated Fusion & 0.479 & 0.480 & 0.438 & 0.440 \\
COCO-Init Gated Fusion & 0.422 & 0.480 & \textbf{0.511} & 0.432 \\
Local-Init MLP Fusion & 0.439 & 0.443 & 0.503 & 0.408 \\
COCO-Init MLP Fusion & 0.427 & 0.384 & 0.469 & 0.381 \\
Local-Init Cross Attention Fusion & 0.408 & 0.333 & 0.436 & 0.312 \\
COCO-Init Cross Attention Fusion & 0.318 & 0.328 & -- & -- \\
\hline
\end{tabular}
\caption{Out of distribution mAP for the small objects across IoU thresholds between 0.5 and 0.9.}
\label{tab:ood_small_objects}
\end{table*}

\subsection{Drone object detection results}
\paragraph{In-Distribution Dataset}
{\it All sized drones evaluation } results for the in‑distribution dataset are presented in Table \ref{tab:in_distribution_results}, covering both the single‑modal RGB Deformable DETR baseline and the multi‑modal architectures, along with their performance under various dropout rates applied in the fusion layers.  Initializing the RGB weights from the local trained Deformable DETR resulted in significantly better results. Among the multi-modal architectures, the gated fusion improved the mAP score by $3.17\%$, followed by the linear fusion which enhanced the mAP score by $2.13\%$. For these two multi-modal architectures, dropout in the fusion layer did not improve the performance. In the case of the MLP fusion, the best mAP is 0.712 and was obtained for dropout rate of $0.1$. The cross‑attention fusion underperformed for all drones mAP metric, primarily because of the insufficient amount of training data.

{\it Small sized drones evaluation:} The largest mAP improvements brought by the acoustic features and fusion layers are observed in the case of the small drones (see Table \ref{tab:small_objects_results}). These are the edge cases that we were seeking to improve in both bounding box and classification score predictions. The gated fusion is the best performer showing $11.1\%$ (dropout rate = 0.1) and $7.2\%$ (dropout rate = 0) improvements when comparing to the RGB deformable DETR architecture. The second best performers are MLP and linear fusion layers. On average, across all dropout rates, the linear fusion was slightly better than the MLP layer whereas the latter obtained the largest mAP of 0.513 for no dropout. The cross attention fusion also revealed good performance for dropout rate of 0.2 on par with the gated fusion when no dropout was utilized. However, the cross attention fusion did not consistently outperform the RGB deformable DETR for all the dropout rates. 

Among the different fusion layers, the gated fusion outperformed the baseline for all dropout rates even when the corresponding RGB weights were initialized from the COCO pretrained version. The linear fusion outperformed all the fusion models initialized with COCO pretrained RGB weights and obtained an mAP result of 0.51 for both dropout rates of $0.1$ and $0.2$.  

{\it Medium sized drones evaluation:} The gated and linear multi‑modal fusion methods provided improvements over the COCO-init RGB baseline (mAP $= 0.765$) for all the dropout rates in the case of the locally initialized Deformable DETR weights (see Table \ref{tab:medium_objects_results}). Local-Init Gated Fusion achieved the best performance with a mAP of $0.799$ (no dropout), representing a $4.4\%$ gain over the baseline. Local Linear Fusion closely followed with $0.797$ (no dropout) and maintained strong scores with dropout $0.1 (0.794)$. Local MLP Fusion also improved upon the RGB baseline, reaching $0.789$ (dropout $0.2$). This architecture was inferior to the baseline for dropout rates of $0$ and $0.3$. Cross‑attention fusion performed moderately ($0.770$ no dropout) but lagged behind the other fusion methods and showed drops at higher dropout rates. 

When initializing the RGB Deformable DETR weights from COCO pretrained version, the gated fusion performed better than baseline in three out of four experiments with different dropout rates. MLP fusion improved over the baseline scores in two out of four dropout experiments. Neither local or cross attention fusion were better than the RGB Deformable DETR baseline. 

{\it Large sized drones evaluation:} For large drones, the RGB baseline already performed strongly (mAP = $0.888$). Local Gated Fusion reached the top performance with a mAP of $0.906$ (no dropout), corresponding to a $2\%$ improvement over the baseline. The Local Gated Fusion achieved higher mAP scores than the baseline across all dropout rate experiments, consistent with the results observed for medium and small drones. This highlights that Local Gated Fusion is the most effective multi‑modal architecture. 
\begin{table*}[t!]
\centering
\begin{tabular}{lcccc}
\hline
\textbf{Method} & \textbf{No dropout} & \textbf{Dropout 0.1} & \textbf{Dropout 0.2} & \textbf{Dropout 0.3} \\
\hline
COCO-Init RGB & 0.759 & -- & -- & -- \\
Local-Init Linear Fusion & 0.807 & 0.788 & 0.779 & 0.797 \\
COCO-Init Linear Fusion & 0.747 & 0.774 & 0.740 & 0.787 \\
Local-Init Gated Fusion & 0.812 & \textbf{0.819} & 0.803 & 0.784 \\
COCO-Init Gated Fusion & 0.745 & 0.760 & 0.754 & 0.770 \\
Local-Init MLP Fusion & 0.783 & 0.756 & 0.805 & 0.747 \\
COCO-Init MLP Fusion & 0.744 & 0.728 & 0.718 & 0.748 \\
Local-Init Cross Attention Fusion & 0.706 & 0.713 & 0.699 & 0.620 \\
COCO-Init Cross Attention Fusion & 0.671 & 0.654 & -- & -- \\
\hline
\end{tabular}
\caption{Out of distribution mAP for the medium objects across IoU thresholds between 0.5 and 0.9.}
\label{tab:ood_medium_objects}
\end{table*}

\begin{table*}[t!]
\centering
\begin{tabular}{lcccc}
\hline
\textbf{Method} & \textbf{No dropout} & \textbf{Dropout 0.1} & \textbf{Dropout 0.2} & \textbf{Dropout 0.3} \\
\hline
COCO-Init RGB & 0.835 & -- & -- & -- \\
Local-Init Linear Fusion & 0.859 & 0.840 & 0.858 & 0.873 \\
COCO-Init Linear Fusion & 0.813 & 0.834 & 0.849 & 0.858 \\
Local-Init Gated Fusion & 0.857 & \textbf{0.880} & 0.864 & 0.858 \\
COCO-Init Gated Fusion & 0.849 & 0.853 & 0.870 & 0.867 \\
Local-Init MLP Fusion & 0.833 & 0.819 & 0.857 & 0.823 \\
COCO-Init MLP Fusion & 0.842 & 0.811 & 0.854 & 0.832 \\
Local-Init Cross Attention Fusion & 0.790 & 0.788 & 0.801 & 0.724 \\
COCO-Init Cross Attention Fusion & 0.736 & 0.806 & -- & -- \\
\hline
\end{tabular}%
\caption{Out of distribution mAP for the large objects across IoU thresholds between 0.5 and 0.9.}
\label{tab:ood_large_objects}
\end{table*}

Local Linear Fusion maintained scores close to or slightly below baseline, with a peak of 0.892 (dropout 0.3). The Local‑Init Linear Fusion method prioritizes the detection of small and medium drones for dropout rates of 0, 0.1, and 0.2. Although it shows a slight decrease in performance for large drones compared to the baseline, overall the linear fusion outperforms the baseline across all drone sizes.   

The Local MLP Fusion performed comparably with the local linear fusion, having slightly decreased mAP scores than the baseline. Cross‑attention fusion trailed behind, with best scores of 0.876 (local) and 0.866 (COCO init) and showed degradation with higher dropout.

\paragraph{Out-of-Distribution Dataset}

{\it All sized drones evaluation - }Table \ref{tab:ood_mAP_results} summarizes the out‑of‑distribution mAP performance across all drone sizes and IoU thresholds between 0.5 and 0.9. The baseline, COCO‑Init RGB, achieves an mAP of $0.741$. All multi‑modal methods are evaluated against this reference to determine how integrating acoustic information and fusion strategies impacts detection performance. 

A clear trend emerges when comparing Local‑Init methods (where RGB Deformable DETR weights were initialized from a locally fine‑tuned model) to COCO‑Init methods (initialized directly from COCO weights) as in the case of the in-distribution dataset results. Across the fusion strategies—linear, gated, and MLP—Local‑Init configurations consistently outperform their COCO‑Init counterparts, highlighting the benefit of transferring a locally adapted RGB representation into the fusion architecture. 

Among all fusion strategies, Local‑Init Gated Fusion stands out as the strongest performer. For a dropout rate of $0.1$, it reaches an mAP of $0.787$, representing a relative improvement of approximately $6.2\%$ over the baseline. This suggests that gated mechanisms are highly effective at selectively combining RGB and acoustic modalities to capture complementary information, particularly when paired with a modest level of regularization.

Local‑Init Linear Fusion also shows strong and stable performance, achieving its best mAP of $0.784$ at a dropout rate of $0.3$. This improvement, approximately $5.8\%$ over the baseline, demonstrates that even a straightforward linear combination of features can meaningfully enhance detection when trained with well‑adapted backbones. Local‑Init MLP Fusion performs competitively, with a peak mAP of $0.778$ at a dropout rate of $0.2$, but it exhibits more fluctuation with respect to dropout settings compared to the gated and linear approaches. This variability suggests that the additional parameters introduced by the MLP fusion layer may require more careful tuning or regularization to reach consistent gains.

In contrast, the COCO‑Init versions of linear, gated, and MLP fusion layers do show improvements over the baseline once dropout is applied, but they do not reach the same level of performance as the Local‑Init counterparts. For example, COCO‑Init Gated Fusion peaks at $0.759$ mAP, only a modest gain over the baseline. Finally, cross‑attention Fusion underperforms in both initialization strategies. 

{\it Small sized drones evaluation -}
The results for small drones reveal that the baseline COCO‑Init RGB model reaches an mAP of $0.443$, which already indicates that detecting small objects under out‑of‑distribution conditions is challenging for the unimodal architecture. When acoustic data and fusion mechanisms are introduced, significant improvements are observed, especially when the RGB backbone is initialized from a locally fine‑tuned model rather than directly from COCO weights.

Among the fusion strategies, the COCO‑Init Gated Fusion stands out as the strongest performer. For a dropout rate of $0.2$, it reaches an mAP of $0.511$, representing an improvement of roughly $15.3$ percent compared to the baseline. Local-Init Gate Fusion showed good performances for $0$ and $0.1$ dropout rates improving over baseline by $9\%$.

The Local‑Init Linear Fusion also performs well, with its highest mAP of $0.503$ obtained at a dropout rate of $0.2$, yielding about a $13.5$ percent improvement over the baseline. Linear fusion remains consistently above the baseline across all dropout settings, indicating stable integration of audio and RGB features. Similarly, Local‑Init MLP Fusion peaks at an mAP of $0.503$ with dropout $0.2$, although its performance fluctuates more depending on the dropout level, which suggests greater sensitivity to hyperparameters. Cross‑Attention Fusion methods, regardless of initialization, do not show meaningful improvements for small drones. 

{\it Medium sized drones evaluation - } The baseline COCO‑Init RGB model reaches a mAP of $0.759$. All the fusion methods surpass this baseline when using locally fine‑tuned RGB weights except the cross-attention. Local‑Init Gated Fusion performs best overall, peaking at $0.819$ for a dropout of $0.1$ representing a $7.9\%$ improvements over the baseline and  remaining strong across other dropout settings. Local‑Init Linear Fusion also shows clear gains, reaching $0.807$ without dropout and maintaining scores above $0.77$ in all configurations. Local‑Init MLP Fusion peaks at $0.805$ with dropout $0.2$, though it is slightly less consistent. 

{\it Large sized drones evaluation - } For large drones, the baseline COCO‑Init RGB model achieves an mAP of $0.835$.
All fusion approaches surpass this baseline except the Cross Attention, with Local‑Init Gated Fusion reaching the highest score of $0.880$ at dropout $0.1$ improving over the baseline by $5.3\%$, showing strong consistency across other dropout values. Local‑Init Linear Fusion also performs well, peaking at $0.873$ with dropout $0.3$ followed by Local‑Init MLP Fusion with mAP score of $0.857$ at dropout 0.2.

COCO‑Init Gated fusion improved over the baseline for all dropout levels.  

\section{Conclusion}

In this paper, we introduced a novel multi-modal WAVE-DETR acoustic and RGB object detector with four different fusion layers combining Wav2Vec2 CNN backbone and Deformable DETR architecture. The gated fusion was the best configuration providing excellent mAP improvements for all sized drones when compared with the single modal RGB Deformable DETR. The most notable improvements of the audio features and fusion architecture was observed in the case of small drones underlying the clear sound signature provided by drones. For this study, we also generated the ARDrone dataset containing more than 7,500 synchronized images and audio segments and they were used to fine tune the WAVE DETR object detector. For future work, we are aiming to test the robustness of the WAVE-DETR architecture against digital and real world adversarial attacks to understand the limitations of our detection system against camouflaged drones. 

\bibliographystyle{IEEEtran}
\bibliography{Main}

\appendix

\subsection{Wav2Vec2 Drone Classification Model}

We leverage Wav2Vec2's low data requirements for the fine-tuning stage to fine-tune a pre-trained Wav2Vec2 model on approximately 80 minutes of audio samples and achieve high-accuracy classification results for the drone classfication task. The in and out of distribution audio datasets were generated in a similar way as the multi-modal ARDrone versions. The audio dataset combined samples from the ARDrone dataset as well as the Drone-Detection dataset  \cite{svanstrom2021real}. 

\subsubsection{Spectral analysis}

We performed a spectral analysis on the in-distribution and out-of-distribution datasets in order to analyze the difference between background and drone audio samples to estimate how well an audio classifier may perform when tasked with differentiating between drone and backround samples. We perform this analysis over the testing datasets. 

\begin{figure}[t]
  \centering
  \includegraphics[width=1\linewidth]{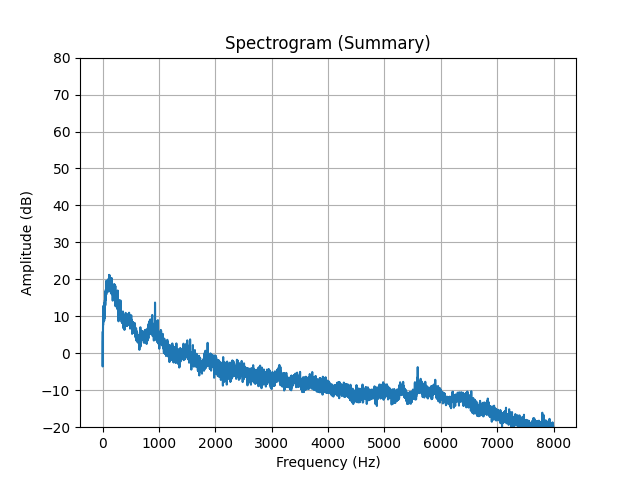}
  \caption{Average spectral analysis of in-distribution test dataset samples with background noise only}
    \label{fig:id_test_avg_spectral_analysis_background}
\end{figure}
\begin{figure}[t]
  \centering
  \includegraphics[width=1\linewidth]{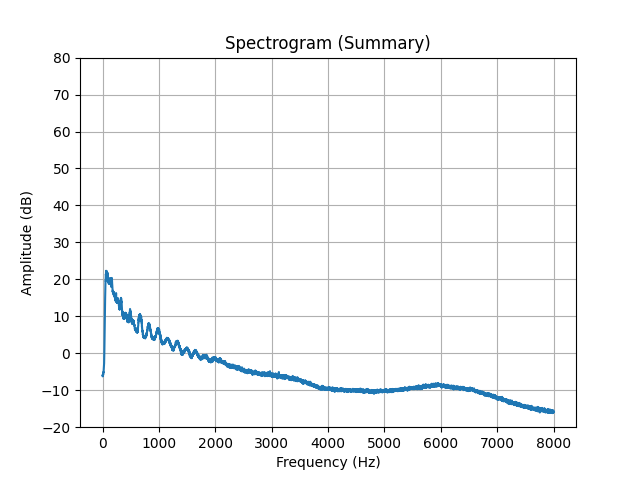}
  \caption{Average spectral analysis of in-distribution test dataset samples with drone hum + background noise}
    \label{fig:id_test_avg_spectral_analysis_drone}
\end{figure}
\begin{figure}[t]
  \centering
  \includegraphics[width=1\linewidth]{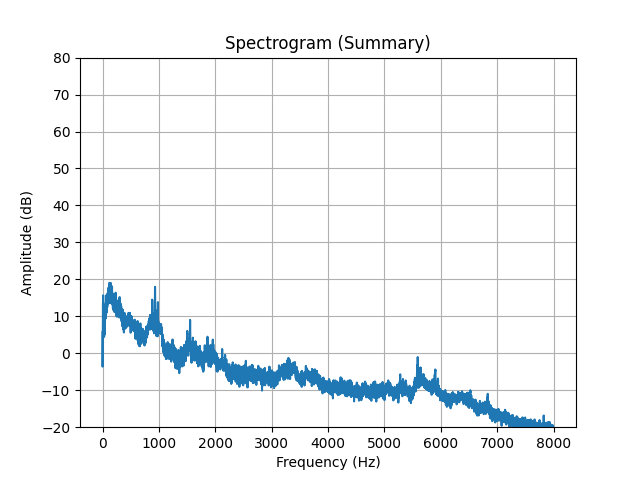}
  \caption{Average spectral analysis of out-of-distribution test dataset samples with background noise only}
    \label{fig:ood_test_avg_spectral_analysis_background}
\end{figure}
\begin{figure}[t]
  \centering
  \includegraphics[width=1\linewidth]{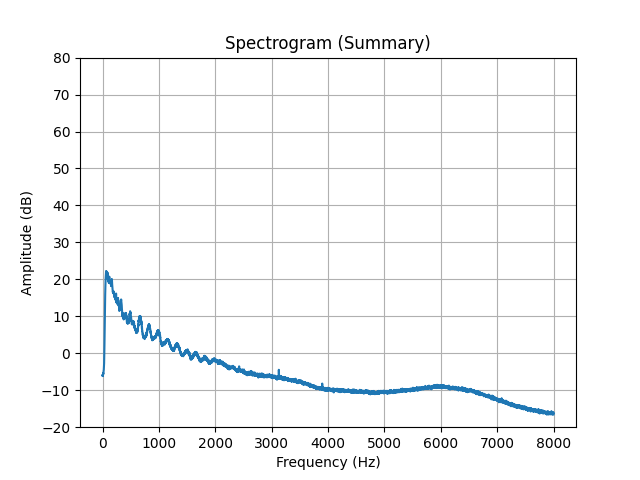}
  \caption{Average spectral analysis of out-of-distribution test dataset samples with drone hum + background noise}
    \label{fig:ood_test_avg_spectral_analysis_drone}
\end{figure}

Figure \ref{fig:id_test_avg_spectral_analysis_background}, and Figure \ref{fig:id_test_avg_spectral_analysis_drone} plot the average amplitude over the individual samples vs. frequency for the in-distribution dataset for the background and drone audio samples in the test datasets. Similarly, Figure \ref{fig:ood_test_avg_spectral_analysis_background}, and Figure \ref{fig:ood_test_avg_spectral_analysis_drone} depict the average amplitude over the individual samples vs. frequency for the out-of-distribution dataset.

As we can see from the amplitude-frequency plots, the amplitude-frequency characteristics of the audio samples consisting of just background noise only vs. the audio samples consisting of the drone hum (in addition to the background noise) exhibit different behavior, especially within the frequency range between 500Hz and 2kHz. As a result, we have confidence that the Wav2Vec2, once fine-tuned for the classification task, should be able to effectively discriminate between audio samples consisting of background noise only and audio samples consisting of the drone hum in addition to the background noise.


\subsubsection{Fine-tuning setup}

We use a pre-trained Wav2Vec2 model as our foundation model for the downstream drone classification task. First we had to re-sample the audio segments to 16kHz to match the sampling rate of the audio samples used for training the foundation model. Next, we concatenated the data from both channels. Since each audio sample has one second in length, we combine the $16,000$ data points present in each channel into a single tensor of $32,000$ data points for each one second  audio sample. Finally, we standardize the concatenated data to create zero-mean, unit-variance tensors. Each tensor is then paired with a label. The label can take on one of two possible classes: Background or Drone.

We fine-tune the model using training and validation dataset sizes of $4,798$ and $1,599$ audio samples, respectively, for the in-distribution dataset and $5,014$ and $1,741$ audio samples, respectively, for the out-of-distribution dataset. We evaluate the models using a test dataset size of $1,600$ audio samples for the in-distribution dataset and $1,744$ audio samples for the out-of-distribution dataset.





\begin{table}[h!]
    \centering
    \begin{tabular}{lcc}
        \hline
        \textbf{Metric} & \textbf{In-Distribution} & \textbf{Out-Distribution} \\
        \hline
        Precision & 0.9961 & 1.0000 \\
        Recall & 1.0000 & 0.9989 \\
        F1-score & 0.9981 & 0.9994 \\
        Balanced Accuracy & 0.9400 & 0.9994 \\
        ROC-AUC & 0.9994 & 1.0000 \\
        PR-AUC & 1.0000 & 1.0000 \\
        MCC & 0.9363 & 0.9894 \\
        \hline
    \end{tabular}
    \caption{Comparison of classification metrics for In-Distribution and Out-Distribution acoustic models.}
    \label{tab:classification_metrics}
\end{table} 

Table \ref{tab:classification_metrics} shows the results of evaluating the two audio classifiers on the test dataset for the in-distribution dataset and out-of-distribution datasets, respectively. PR-AUC stands for precision-recall area under the curve and Matthews Correlation Coefficient (MCC) was also given. As the results show, both models exhibit excellent performance in successfully discriminating between audio samples containing only background noise and audio samples containing a drone hum in addition to the background noise.

With the results demonstrated in Table \ref{tab:classification_metrics}, it may be tempting to simply use the audio classifier for effectively performing drone detection. However, using audio for drone detection is limited in a few ways. First, audio captured with a single microphone has very limited information for ascertaining the spatial positioning of any drone, thus limiting the countermeasures that could be employed against any detected drone. Second, using data from a single modality introduces a single point of failure, where various attack measures (e.g., audio-based camouflage, etc.) or even non-ideal environmental conditions (e.g., loud sounds, etc.) could negatively affect the performance of the audio classifier.  

\end{document}